\documentclass[11pt,a4paper]{article}
\usepackage[hyperref]{acl2019}
\usepackage{times}
\usepackage{latexsym}
\usepackage{bm}
\usepackage{multirow}
\usepackage{amsfonts}
\usepackage{amsmath}
\usepackage{booktabs}
\usepackage{xspace}
\usepackage{url}
\usepackage{graphicx}

\aclfinalcopy 

\title{An end-to-end Neural Network Framework for Text Clustering}

\author{Jie Zhou \\
  Pattern Recognition Center,    \\
  WeChat AI,Tencent Inc. \\
  \texttt{withtomzhou@tencent.com~~} \\\And
  Xingyi Cheng \\
  \\
  \\
  \texttt{derrickzy@gmail.com} \\\And
  Jinchao Zhang \\
  Pattern Recognition Center \\
  WeChat AI,Tencent Inc. \\
  \texttt{~~dayerzhang@tencent.com}
  }

\date{}

\begin{document}
\maketitle
\begin{abstract}
  The unsupervised text clustering is one of the major tasks in natural language processing (NLP) and remains a difficult and complex problem. Conventional \mbox{methods} generally treat this task using separated steps, including text representation learning and clustering the representations. As an improvement, neural methods have also been introduced for continuous representation learning to address the sparsity problem. However, the multi-step process still deviates from the unified optimization target. Especially the second step of cluster is generally performed with conventional methods such as k-Means.   We propose a pure neural framework for text clustering in an end-to-end manner. It jointly learns the text representation and the clustering model. Our model works well when the context can be obtained, which is nearly always the case in the field of NLP.  We have our method \mbox{evaluated} on two widely used benchmarks: IMDB movie reviews for sentiment classification and $20$-Newsgroup for topic categorization. Despite its simplicity, experiments show the model outperforms previous clustering methods by a large margin. Furthermore, the model is also verified on English wiki dataset as a large corpus.
\end{abstract}

\section{Introduction}

The knowledge of text categorization benefits multiple natural language understanding tasks, such as dialogue~\cite{Ge-Xu-ACL2015}, question-answering~\cite{Yao-Durme-ACL14}, document summarization~\cite{Bairi-Bilmes-ACL2015}  and information retrieval~\cite{manning2008introduction}. Supervised methods for text classification are often applied in a wide range and generally perform better than unsupervised clustering methods. However, with the explosive growth of the Internet, unsupervised methods begin to reveal its advantages.

Labeling the text data costs heavy manual efforts. It is impractical to afford such a cost for a large amount of data. This problem is seriously enlarged when we don't have the prior information of the corpus. After processing more data, we will be aware that the total number of categories might be increased, or the boundaries between different categories should be re-defined. People might also have additional interests to look at the effects of a various number of categories at different levels on the system. All this causes the much more increased labeling efforts.




Clustering methods circumvent the above difficulties since they do  not require data annotations. But the following difficulties are still in front of us, especially in text clustering. First, the large vocabulary brings sparsity problem in text representations, while conventional clustering tools such as k-Means are mainly designed for dense features. Second, the exploded corpus size and the increased requirements on the number of categories decrease the efficiency of conventional tools further. Third, conventional methods often suffer from the isolated stages for learning the text representation and training the clustering model, which leads to the difficulty in unified optimization.

Neural methods address the sparsity problem by representing the text with continuous distributed vectors (embeddings)~\cite{Le-Mikolov-ICML2014,kiros2015skip,tai2015improved,triantafillou2016towards}. But the clustering step still relies on conventional tools such as k-Means \cite{manning2008introduction}, that the experience of neural methods in handling big data has little contribution to speeding up the whole pipeline.


In this paper, we propose an end-to-end neural framework for text clustering. Instead of conventionally trying to find out which cluster each instance belongs to, our model clusters the instances by determining whether two instances have the same or different categories, which is a binary classification problem. The true category distribution is considered as a latent variable and represented by a hidden layer vector in the neural framework. The binary label is obtained by a simple artificial rule. Implemented with a pure neural network, our framework unifies the representation learning and clustering procedures into an end-to-end system.

We evaluate our method on two widely used benchmarks: IMDB Movie Reviews (IMDB) \cite{Maas-Potts-ACL2011} for sentiment classification and 20-Newsgroup (20NG) \cite{Lang-newsweeder-ICML95} for topic categorization. Experimental results show that our method outperforms conventional methods by a large margin. We also verify the performance of our model on the English wiki corpus which has neither predefined categories nor clear boundaries between categories.

\section{Method}

Most clustering methods try to find which cluster the current sample belongs to in an iterative way, explicitly in real feature value space such as k-Means~\cite{lloyd1982least, manning2008introduction}, or implicitly in parameter space such as Gaussian Mixture Model~(GMM)~\cite{jain2010data} and Latent Dirichlet Allocation~(LDA)~\cite{Blei-Jordan-JMLR2003}.

Instead, ``whether these two samples belong to the same cluster'' conducts our model optimization. Category information is not the final output of our model. Instead, it is learned as an latent variable, a hidden layer in the neural network framework.  Thus, the clustering problem is transformed into a binary classification problem. The binary labels are automatically constructed under the following rule which is widely applicable.

In clustering stage, a sample refers to a single sentence, or a word sequence more generally. In the inference stage, we can obtain the category information at the sentence level, or any higher level by averaging the category distribution over each sentence within this level (e.g. paragraph or document).

The spirit of utilizing a pair of instances comes from the field of learning to rank~\cite{cao2007learning, liu2009learning} and is also used in image classification or area~\cite{koch2015siamese}. Noted that previous works requires manual label while our work is purely an unsupervised learning.

\subsection{Prerequisite and Pseudo Labels}
\label{sec:Method-PreAss}

A pair of word sequences is built as an input instance in our neural end-to-end text clustering model. It is natural that the co-occurrence of sequences within a short distance is likely to put them in the same category (label $1$) than those far from each other (label $0$). We will not pay attention to the precision of these pseudo labels. Our main goal is to exhibit that the model will finally converge to give the true categories against these noisy pseudo labels.

Two points will be further explained in the following.
First, in the actual situation, two sequences distant from each other may also have the same category, and those next to each other may have different categories. The detailed inconsistency will not affect our model performance. We only need this assumption statistically established, since our neural network framework is also a statistical model resistant to high level label noise as we tested in experiments part. This result is out of our expectation at the beginning but quantitatively verified.

Second, our method is not restricted by the text structure. The text corpus could be organized at paragraph level, document level, or without any structural boundaries. Our method holds as long as the distance can be defined.  Even under a specific condition where all sentences are isolated without context information, a sub-sequence could be considered as a sample.  Then an instance, a pair of two samples, within the same sentence have positive label.

\subsection{Instances Construction}

\begin{figure}[!htb]
	\begin{center}
		\includegraphics[angle=0,width=0.4\paperwidth]{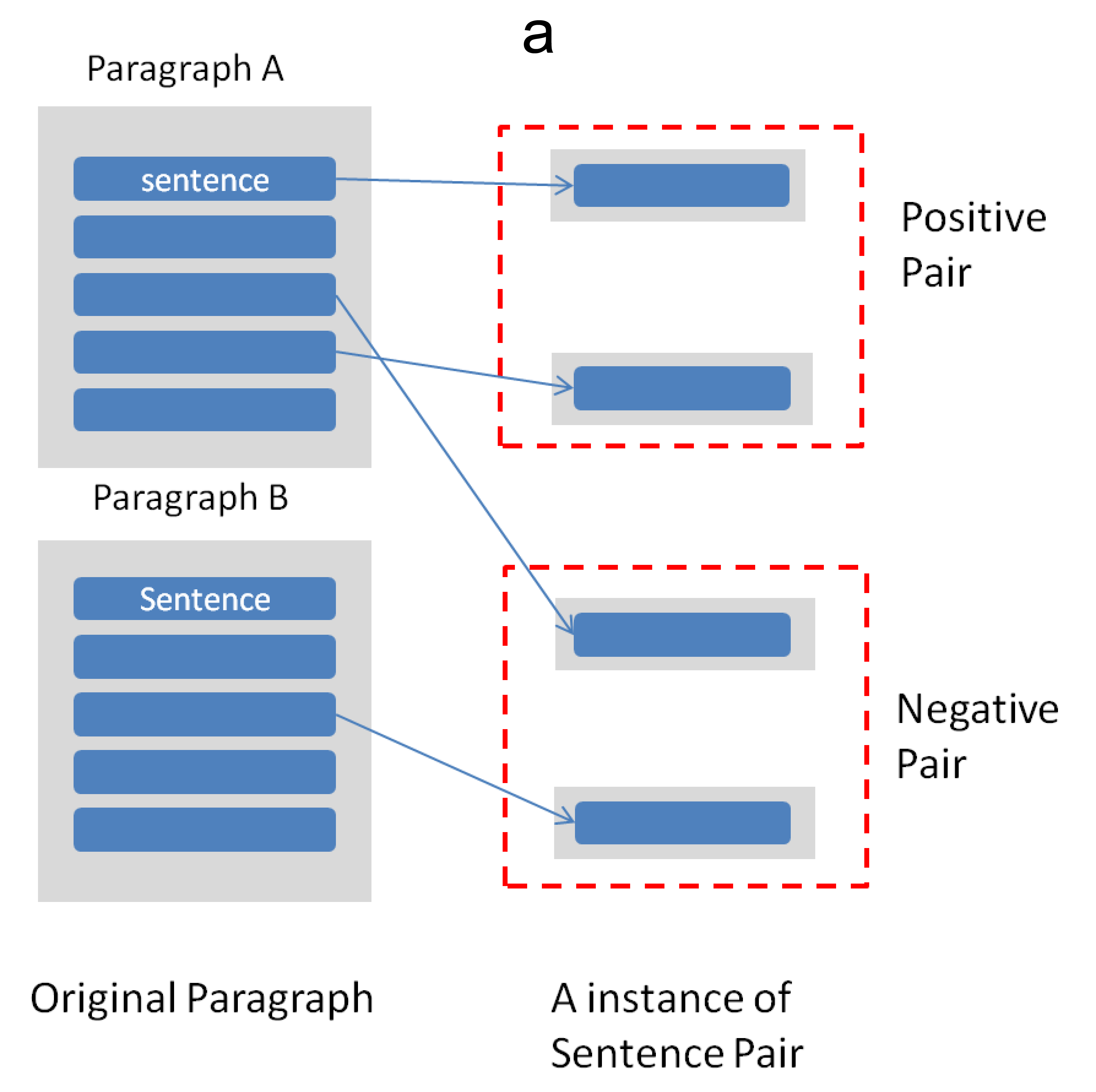}
		\includegraphics[angle=0,width=0.4\paperwidth]{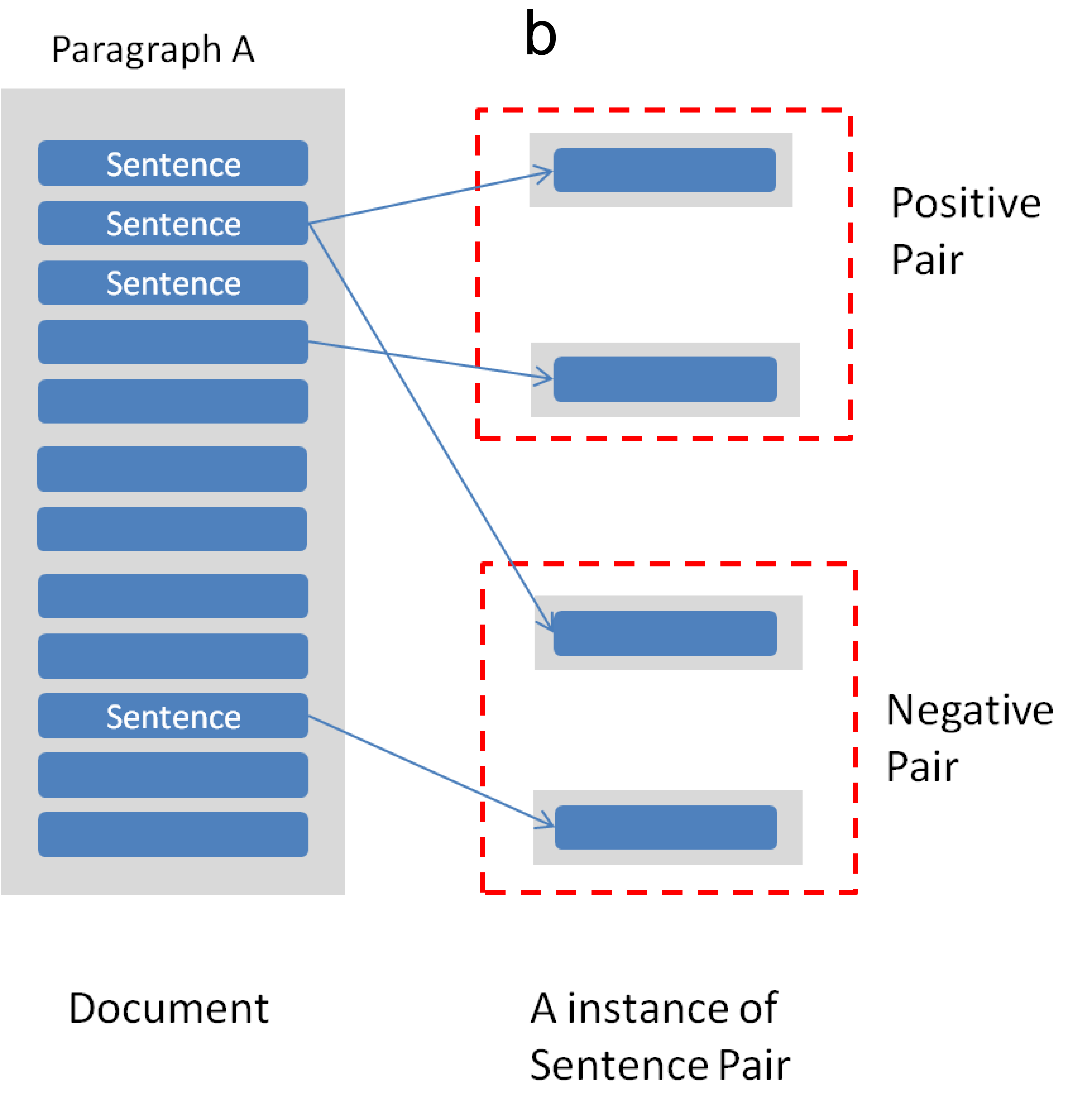}
		\caption{Construct instances from a document with (top graph a) or without structural boundary (bottom graph b).}
		\label{fig:Method-PreAss-Ins}
	\end{center}
\end{figure}

Examples of positive and negative instances are given in Fig.~\ref{fig:Method-PreAss-Ins}. For the corpus organized at paragraph level, we randomly select two sentences from the same or different paragraphs to build a positive (within the same category) or negative (with different categories) instance respectively (Fig.~\ref{fig:Method-PreAss-Ins}a). For the corpus without any structural boundary, we select two sentences next to each other to build a positive instance and those far from each other to build a negative instance (Fig.~\ref{fig:Method-PreAss-Ins}b).

In inference stage, we predict the category distribution for each sequence. The higher level (such as paragraph) category is obtained by averaging over all sequences within this level.

\subsection{Model Topology}
\label{sec:Model-Top}

\begin{figure}[!htb]
	\begin{center}
		\includegraphics[angle=0,width=1.0\columnwidth]{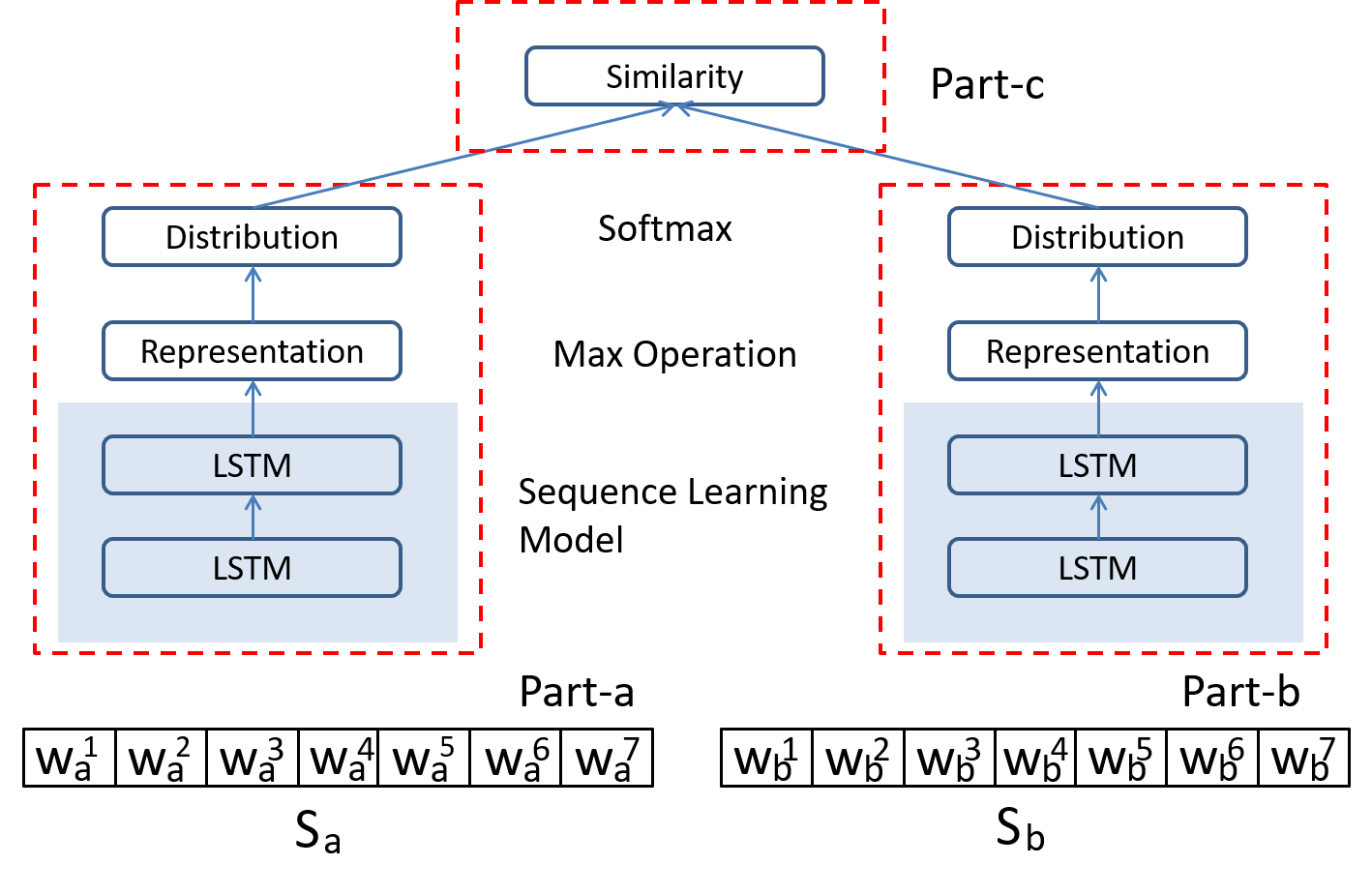}
		\caption{The end-to-end neural network topology for text clustering.}
		\label{fig:method-model-unsup}
	\end{center}
\end{figure}

The whole framework of our end-to-end clustering system is shown in Fig.~\ref{fig:method-model-unsup} which includes three parts.

\begin{itemize}
	\item \textbf{Part-a} and \textbf{Part-b} deal with the two input sequences in a pair respectively. They share the same topology and parameters. The words in input sequences are mapped to 300-dimensional GloVe word embeddings trained with 840 billion tokens \cite{pennington2014glove} and fed to the stacked recurrent neural network to generate the representation of input sequence. The \emph{softmax} layer generates the probability distribution of categorization. We only need to set the category size to the distribution vector of the \emph{softmax} layer.
	\item \textbf{Part-c} measures whether the two sequences belong to the same category using \emph{cosine} similarity of the two distribution vectors generated by part-a and part-b.
\end{itemize}

We employ the bi-directional long short-term memory (LSTM) \cite{Hochreiter-Schmidhuber-NC1997} network to process the input sequences (\{$w_a^i$\} and \{$w_b^i$\}). Our framework is flexible to use different sequence learning layers, for example, using stacked LSTMs or CNNs. The $\textrm{Max}(\cdot)$ operation is employed to extract the representation after LSTM layers. $\textrm{Max}(\cdot)$ means to assign the maximum value from a time series $i$ of input vectors of each dimension  to the output. Then a \emph{softmax} function is used to transform the representation into a normalized vector. \emph{cosine} metric is used to measure the similarity between two normalized vectors. At last, we compute the square error (SE) cost between the \emph{cosine} similarity and the pseudo label $l_{a,b}$ ($1$ for positive pair $\{a,b\}$ and $0$ for negative one).

Although it is not mathematically or strictly guaranteed that the normalized vector represents the category distribution, we unseal this phenomenon in our experiments.

In the inference stage, only part-a is used and it works like a classifier to predict the category distribution of the input sequence.

The detailed parameter setting will be shown in experiments part. We only stress here that the dimension of normalized vector (category distribution) can be arbitrarily assigned, which denotes the number of categories. Since we verified our methods on two benchmarks with the knowledge of true labels, we assign the true categories as the dimension of the distribution layer. This is consistent with conventional pipelines.

\subsection{Further thinking}
\label{sec:Method-Difficulty}

\begin{itemize}
\item
	\textbf{on Label: noise difficulty}: Our strategy to construct the pseudo labels will inevitably  bring much noise into the training instances. In the worst case of two-category clustering problem with homogeneous category distribution, the labels of negative pairs are pure noises. Because half of the instances are really from different categories while the other half are from the same category.
\item
	\textbf{on Prediction: contradiction difficulty}: At the beginning stage in training, instances from the same category may activate different softmax dimensions. It means different softmax indexes may denote the same category while several other categories activate the same softmax dimension. The following training stage has to solve this contradiction. This contradiction also leads to the iterative instability.
\end{itemize}

Corresponding to these two difficulties respectively, experiments on two typical benchmarks will be analyzed to provide a further insight.

\section{Experiments}

We resort to ground truth to evaluate our clustering method in a quantitative and rigorous way. We choose two widely used benchmarks: IMDB movie reviews (IMDB) dataset and 20-Newsgroup (20NG) dataset. On IMDB, which is a two-category problem, we will test our ability of the resistance to strong pseudo label noise in the assumed negative pairs (Noise difficulty). On 20NG, we will test the model ability in dealing with multi-category problems (Contradiction difficulty). For further verification, we also select the English Wiki (EnWiki) dataset, which consists of huge data and does not have clear boundaries between categories as in IMDB and 20NG.

A series of evaluation methods are used in our experiments. Metrics using ground truth are Accuracy, F-score (weighted, micro and macro), Adjusted Random Index(ARI), Adjusted Mutual Information(AMI) and Normalized Mutual Information(NMI) \cite{vinh2010information}.  We also employed the internal metric Davies-Bouldin Index (DBI) \cite{davies1979cluster} which does not rely on ground truth.

After obtaining the clustering results, we follow the general way to use Hungarian algorithm \cite{Papadimitriou-Steiglitz-1982} to assign the predicted category name to each cluster for evaluation. The Hungarian algorithm searches the mapping of category name to each cluster with highest accuracy score from all possible category permutations.

\subsection{IMDB}

\subsubsection{Dataset}

IMDB \cite{Maas-Potts-ACL2011} is one of the largest open datasets for sentiment analysis and is generally used as a two-category classification benchmark. Each paragraph is considered as a single review which consists of several sentences.  This dataset has three partitions: $25$k labeled reviews for training, $25$k labeled reviews for testing and $50$k unlabeled reviews. There are two types of labels and the label distributions in training and testing data are balanced. We combine the original training part and unlabeled part to form our training set. We ignore the label information when training our clustering model. The performance is evaluated on the original test set.

\subsubsection{Model Training}
\label{sec:exp-imdb-mod}

We prepare the instances for our model training as introduced in the above instances construction section. Each positive pair is randomly chosen from the same paragraph and each negative pair is built from different paragraphs. We have equal numbers of both sets. Since IMDB is a two-category problem, we meet with the noise problem in sampling the negative instances as mentioned in previous sections. According to our sampling rule, half of the negative instances are correctly labeled and the other half are wrongly labeled. Thus, the negative instances are pure noise. Nevertheless, our positive instances are guaranteed to be correctly labeled.

For the IMDB dataset, we use the single layer bi-directional LSTM to process the input sequence. The LSTM layer has $256$ memory blocks. The learning rate is set to be $1\times10^{-3}$ and $L_2$ regularization is set to be $1\times10^{-4}$. The softmax layer dimension is equal to the number of categories which is $2$ in this task.

\subsubsection{Results}

We show our clustering results together with several conventional methods in Tab.~\ref{tab:exp-imdb-res}. The first two methods are based on k-Means algorithm. Singular value decomposition (SVD) or Paragraph Vector (PV) are employed to obtain low dimensional vectors to represent the text and then cluster these vectors using k-Means. The vector dimension is set to be $128$.  People also use a topic model Latent Dirichlet Allocation (LDA) to carry out this task \cite{Maas-Potts-ACL2011} based on sparse text representations.

Our neural based method outperforms the others by a large margin of nearly $6$ points. We find this accuracy is not far from a simple supervised method MNB-uni (Multinomial Naive Bayes with uni-gram) which gives $83.6$ \cite{Wang-Manning-ACL2012}.

\begin{table}[h]
	\small
	\begin{center}
		\begin{tabular}{|l|l|}
			\hline \bf Approach & \bf Acc  \\ \hline \hline
			SVD+k-Means & 62.9  \\
			PV + k-Means  & 72.3  \\
			LDA  & 67.4   \\
			NMF &  62.3 \\
			\hline
			Ours   & 78.1   \\
			\hline
		\end{tabular}
	\end{center}
	\caption{\label{tab:exp-imdb-res} Text clustering results on IMDB dataset. We compare our neural based method with four conventional methods, SVD+k-Means, PV(Paragraph Vector)$+$k-Means \cite{pelaez2015sentiment}, LDA \cite{Maas-Potts-ACL2011} and NMF (Non-negative matrix factoring). }
\end{table}

\subsubsection{Analysis}

As we mentioned in previous section, there are a lot of contradiction updates and instabilities during the model training. Especially the strong noises in this two-category problem strengthen this obstacle.

\begin{figure}[!htb]
	\begin{center}
		\includegraphics[angle=0,width=0.9\columnwidth]{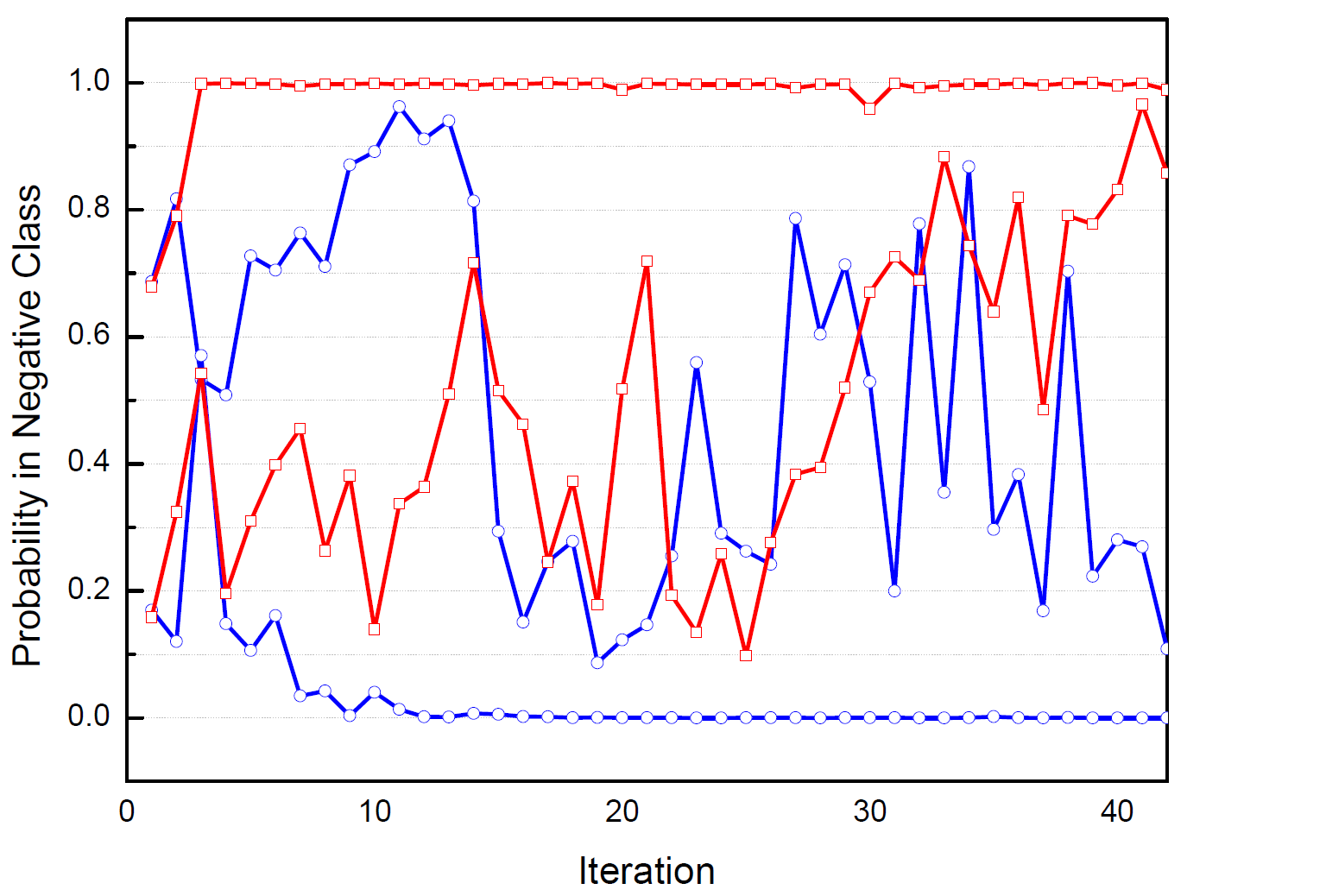}
		\caption{The evolution of the probability in negative class.}
		\label{fig:exp-imdb-process}
	\end{center}
\end{figure}

We randomly initialize the network, and all instances are predicted randomly at the beginning. When a negative pair is predicted to be a positive pair, both input sequences are inclined to be moved into the other classes. This phenomenon results in the above difficulties. This process is shown in Fig.~\ref{fig:exp-imdb-process}. We exhibit the update process of $4$ selected instances. Two negative pairs depicted with blue lines (circle) and two positive pairs depicted with red lines (square). Both positive and negative pairs include one easy instance and one hard instance respectively. Easy instance denotes the instance that converges fast into its true state and stays at its state, as the top flat curve and the bottom flat curve behave in Fig.~\ref{fig:exp-imdb-process}. Hard instances, the two middle lines in Fig.~\ref{fig:exp-imdb-process}, fluctuate dramatically between two states (several times), and then converge to their final states.

In Fig.~\ref{fig:exp-imdb-process}, an instance that belongs to one class is first assigned to the other class and then moves back to its true class. Our model exhibits its ability to overcome local minimums. On the contrary, k-Means based methods often restricted by it greedy properties. Once an instance is assigned to one class, it is very difficult to escape from this class.

\subsection{20-Newsgroup}

\subsubsection{Dataset}
The 20-Newsgroup (20NG) dataset \footnote{http://qwone.com/~jason/20Newsgroups/} \cite{Lang-newsweeder-ICML95} is a widely used benchmark for multi-category document clustering. It contains $18,846$ documents across $20$ different categories. The dataset is split into train set and test set with $11.3$k  and $7.5$k documents respectively.  These $20$ categories are organized into $6$ main subjects as listed in Tab.~\ref{tab:exp-20NG-sub}. Categories within the same subject are closely related to each other, and the others are highly partitioned. Due to its difficulty, a lot of works focus on a selected subset of categories or a group of selected category pairs. In our work, we addressed this problem on all $20$ categories.

\begin{table}[h]
	\small
	\begin{center}
		\begin{tabular}{|l|l|}
			\hline
			comp.graphics & rec.autos \\
			comp.os.ms-windows.misc & rec.motorcycles \\
			comp.sys.mac.hardware & rec.sport.baseball \\
			comp.sys.ibm.pc.hardware & rec.sport.hockey \\
			comp.windows.x & \\
			 \hline
			 sci.crypt & talk.religion.misc \\
			 sci.electronics & alt.atheism \\
			 sci.med & soc.religion.christian \\
			 sci.space & \\
			\hline
			misc.forsale & talk.politics.misc \\
			 & talk.politics.guns \\
			 & talk.politics.mideast \\
			\hline
		\end{tabular}
	\end{center}
	\caption{\label{tab:exp-20NG-sub} Two-level categories of 20-Newsgroup.}
\end{table}

For a better illustration, we also provide experiment results on another partition with selected $4$ groups of categories \cite{Zhang-Si-AIGIR2011}. The $4$ group names are `comp', `sci', `rec', `talk'. The first word of each category name denotes the group it belongs to. Then the subset will be clustered into $4$ classes.

With these two experiments, we will compare the performance at both high and detailed levels of category partition. Noting that on the contrary to the experiments on IMDB dataset, here we have at most $20$ homogeneous classes. This means almost $95\%$ negative instances (pairs of sentences) have correct assumed labels.

\subsubsection{Model Training}

We follow the same way as used in IMDB experiment to prepare the instances. We construct the positive and negative instances by sampling the sentence pairs  from the same paragraph or different paragraphs respectively. The final training corpus consists of $20\%$ positive instances and $80\%$ negative instances.

We can set an arbitrary cluster number (dimension of softmax layer) to our model but this will introduce post processing steps for evaluation. For the sake of simplicity, we set the cluster number to be equivalent to the ground truth.

For the 20NG dataset, two stacked bi-directional LSTMs are used to process the input sequence, as shown in Fig.~\ref{fig:method-model-unsup}. In clustering the whole $20$ categories, a smaller learning rate $l_p=1\times10^{-4}$ is used. While on a selected subset with $4$ categories, we keep the same learning rate $l_p=1\times10^{-3}$ as in IMDB experiments.

\begin{figure}[!htb]
	\begin{center}
		\includegraphics[angle=0,width=0.9\columnwidth]{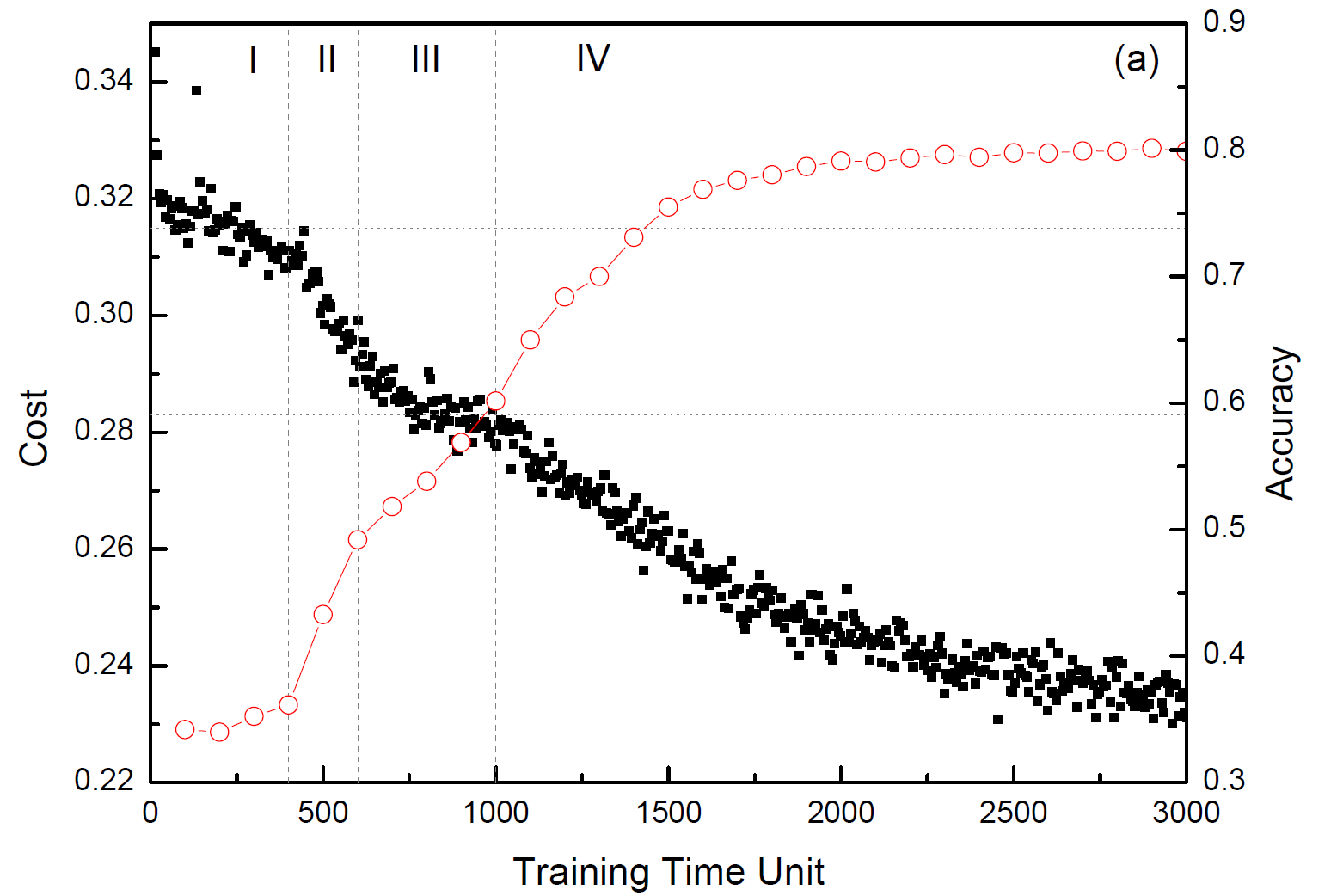}
		\includegraphics[angle=0,width=0.86\columnwidth]{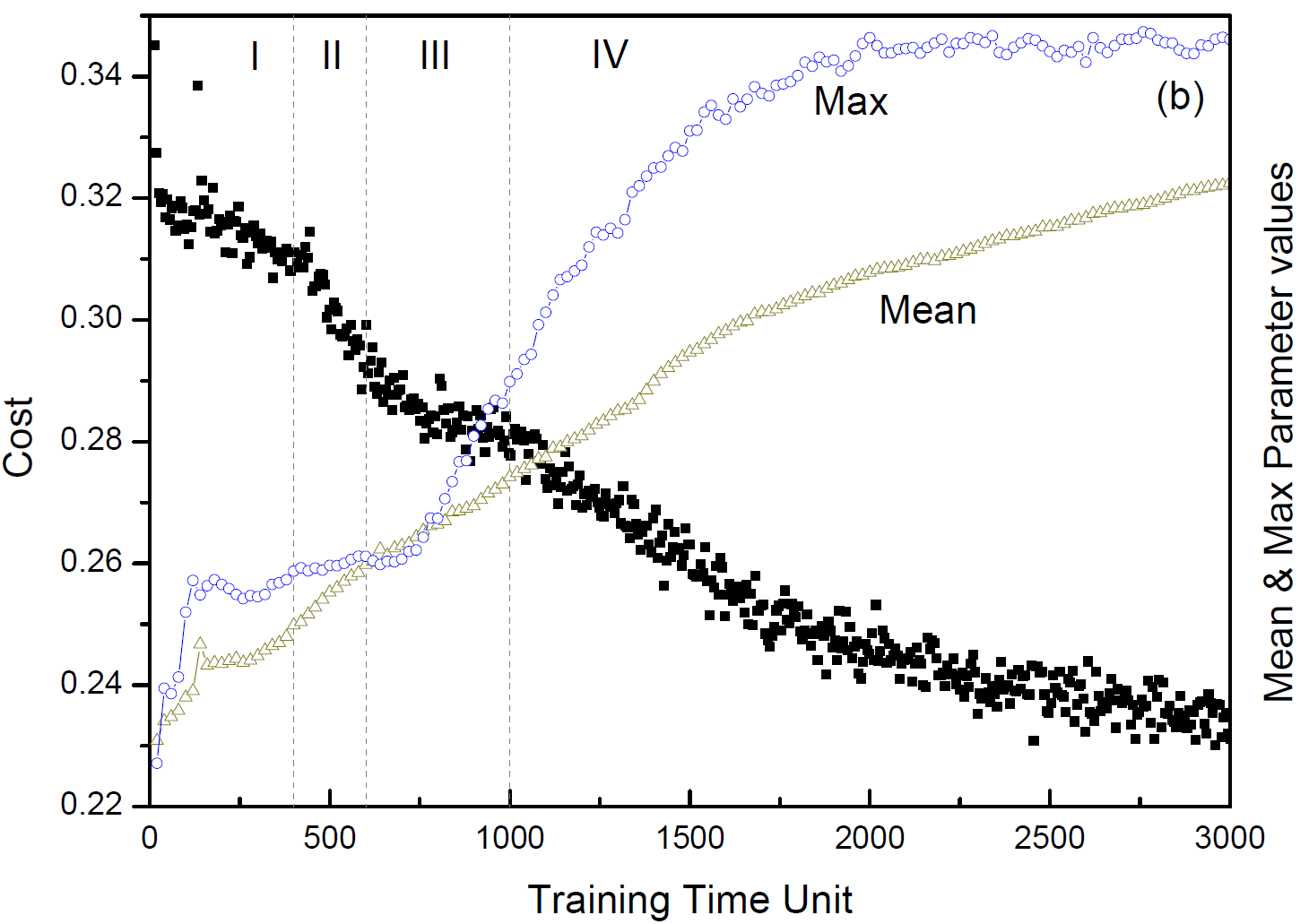}
		\includegraphics[angle=0,width=0.9\columnwidth]{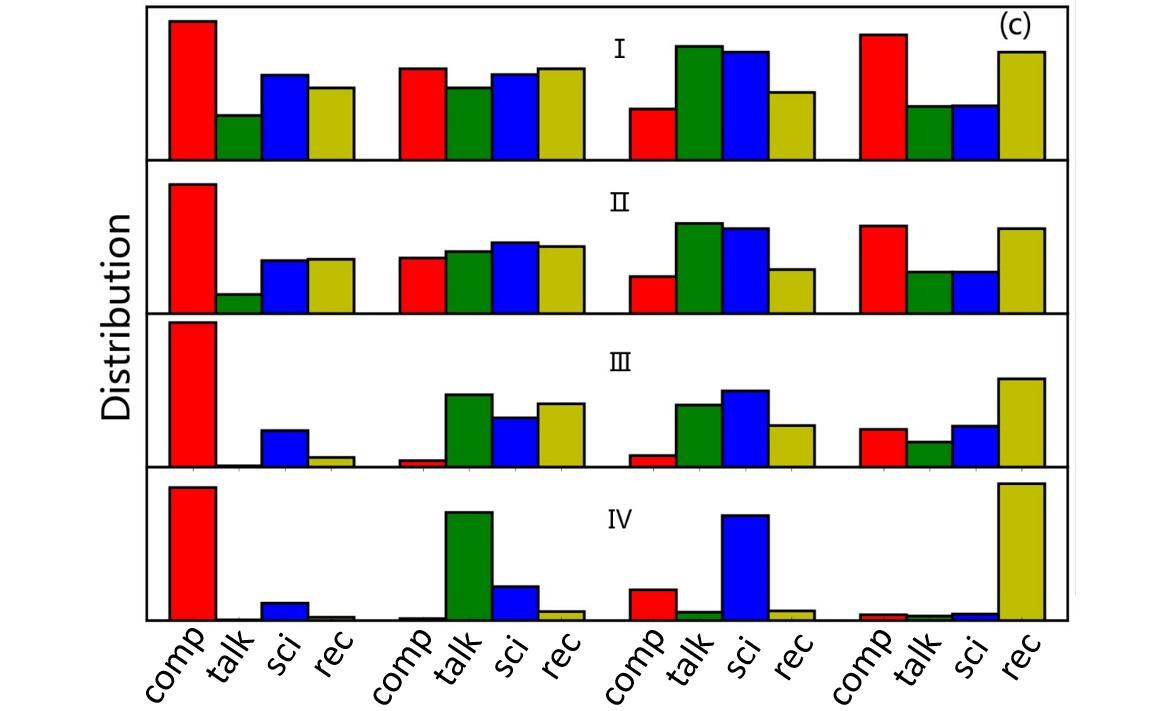}
		\caption{(a): Black line: evolution of the training cost value. Red line: the clustering accuracy. The procedure can be split into $4$ stages. (b): evolution of the mean and max value of LSTM layers. (c): The distribution of $4$ types of true labels in each cluster at $4$ stages respectively.}
		\label{fig:exp-20ng-evol}
	\end{center}
\end{figure}

\begin{table*}[t!]
	\small
	\begin{center}
		\begin{tabular}{|l|l|l|l|l|l|l|l|}
			\hline \bf Approach &F-score & F-score-micro & F-score-macro& Accuracy & ARI & AMI & NMI \\
			\hline\hline
			$20$ categories & &&&&&&\\
			\hline
			TF-IDF+k-Means     &  33.0 & 33.1 & 32.0 & 33.1 & 14.2 & 33.7 & 37.0 \\
			NMF         &  35.1 & 34.2 & 34.2 & 34.2 & 18.3 & 33.4 & 34.5 \\
			LDA         &  37.5 & 31.6 & 30.3 & 31.6 & 13.8 & 33.8 & 37.1  \\
			LSA+k-Means &  39.4 & 37.5 & 38.1 & 37.5 & 18.1 & 37.0 & 38.7 \\
			\hline
			Ours & \bf 42.2 & \bf 50.8 & \bf 40.6 & \bf 50.8 & \bf 41.6 & \bf 53.1 & \bf 57.1 \\
			\hline\hline
			$4$ categories & &&&&&&\\
			\hline
			TF-IDF+k-Means & 59.8 & 58.4 & 59.3 & 58.4 & 25.6 & 29.1 & 29.2 \\
			NMF          & - & - & - & 64.3 & - & - & 44.3   \\
			LDA          & 54.7 & 61.8 & 53.2 & 61.8 & 35.3 & 34.3  & 37.7  \\
			LSA+k-Means  & 63.7 & 63.1 & 63.3 & 63.1 & 28.7 & 33.5  & 34.7   \\
			GMM         & - & - & - &  51.9 & - & - & 20.5  \\
			PLSA        & - & - & - &  66.5 & - & - & 47.6 \\
			M$^2$DCU    & - & - & - &  69.0 & - & - & 40.8 \\
			\hline
			Ours & \bf 78.9 & \bf 79.1 & \bf 78.6 & \bf 79.1 & \bf 55.3 & \bf 52.9 & \bf 53.0 \\
			\hline
		\end{tabular}
	\end{center}
	\caption{\label{tab:exp-20ng-res-20} The clustering results on $20$ categories and $4$ categories.  We compare our method with TF-IDF+k-Means, NMF, LDA, LSA+k-Means, GMM, PLSA and M$^2$DCU \cite{Zhang-Si-AIGIR2011}.  We obtain the best performance on both category partitions with all evaluation metrics.}
\end{table*}

\subsubsection{Results}

First, we cluster the full 20NG dataset into $20$ categories. This is the most difficult partition on this benchmark, because of the number of categories and the high similarity between pairs of similar categories, such as ``rec.sport.baseball'' and ``rec.sport.hockey" (see Tab.~\ref{tab:exp-20NG-sub}).

We compare our results with widely used baselines, including nonnegative matrix factorization (NMF), latent dirichlet allocation (LDA), Latent semantic analysis (LSA)+k-Means and TF-IDF+k-Means, gaussian mixture model (GMM) and probabilistic latent semantic analysis (PLSA). There are limited comparable other works on $20$ category partion. Many of people focus on a subset of this problem, including clustering a group of selected categories, or a pair of categories. In Chen $et~al.$ \shortcite{Chen-Xioing-AAAI2016} and Palla $et~ al.$  \shortcite{Palla-Knowles-NIPS2012}, a full version of this data set is investigated with  dirichlet process based methods (MMDPM and DPVC), but only a smaller vocabulary of $250$ words is used with the consideration of efficiency. Thus results obtained therein (with f-score of $10.0$) are much lower than ours (see Tab.~\ref{tab:exp-20ng-res-20}). On $4$ category partition, we also have the results in \cite{Zhang-Si-AIGIR2011} for comparison.

With $20$ categories, we obtain the best performance with all evaluation metrics.
We note that the improvement on f-score is smaller than that on ARI and accuracy. During the training, we set the number of clusters to be $20$. Actually our model only predicts $17$ classes, that is no instance is predicted to be the other $3$ classes. This phenomenon decreases the f-score much, while ARI is designed to specifically deal with this problem. Accuracy is basically an instance level evaluation rather than cluster level. So the improvement lies between that of F-score and ARI. A larger cluster size could be set for better evaluation score. But this will introduce some post processing techniques. Here we just show a straightforward way in model training which has demonstrated its advantages over other works.

Next, for a better illustration of the clustering performance, we only consider the selected $4$ groups of categories ($4$-category simply speaking), which are `comp', `rec', `sci' and `talk'. Here we use accuracy as the evaluation metrics in accordance with the work of Zhang $et~al.$ \shortcite{Zhang-Si-AIGIR2011}. We list our results together with conventional tools, such as  k-Means, GMM, PLSA, and Max Margin Document Clustering with Universum (M$^2$DCU)\cite{Zhang-Si-AIGIR2011}  results in Tab.~\ref{tab:exp-20ng-res-20}. Experiment results show that our method has the best performance with all evaluation metrics. Compared to $20$-category results, here we obtain the consistent improvement amplitude on Accuracy, ARI and F-score because all $4$ categories are predicted.

After considering performance on both category levels and looking into the detailed cases, we find our model works well in predicting the high level categories. Mistakes exist in distinguishing the subtle differences, such as ``rec.sport.baseball'' and ``rec.sport.hockey", ``talk.politics.guns'' and `talk.politics.mideast'', where the improvement is also enlarged with our model. In addition, compared to the improvement on IMDB benchmark, it appears that our model is more advantageous under difficult conditions.

\begin{table*}[ht!]
	\small
	\begin{center}
		\begin{tabular}{|l|l|}
			\hline
			cluster-1  & He went $1$-$3$ with a $8.16$ era in $32$ innings pitched. \\
			           & The cardinals responded by scoring three runs in the bottom of the fourth inning. \\
			           & Rangers won the match 3-0 and therefore won the title. \\
			\hline
			cluster-2 & Religious believers may or may not accept such symbolic interpretations. \\
			          & Opposing views are not non-existent within the realm of christian eschatology. \\
			          & Many great philosophers have spoken of the importance of exercising both humility and confidence. \\
			\hline
			cluster-3 & Chrysler corporation only made 701 gtx convertibles in 1969. \\
			          & Cosworth technology was then renamed as mahle powertrain on 1 july 2005. \\
			          & In $2009$ the route gained five new alexander dennis enviro$200$ diesel-electric hybrid single-deckers. \\
			\hline
			cluster-4 & T-bag responds by starting to poison lechero's mind against his men. \\
			          & When he refuses, she slams the door on him in apparent disgust. \\
			          & Later that night, buffy gets drunk with spike at his crypt. \\
			\hline
			cluster-5 & Unlike all other final fantasy games, players cannot manually equip characters with armor. \\
			          & Produced by bandai, the game was first introduced in Japan in February 2003. \\
			          & Various weapons and accessories can be attached to many player and ai objects. \\
			\hline
			cluster-6 & In april 1944, the squadron shifted from bomber escort to ground attack duties. \\
			          & The entire squadron then transferred to tunis in June to attack enemy shipping. \\
			          & On 6 March 1945, the two gunboats arrived at eniwetok. \\
			\hline
		\end{tabular}
	\end{center}
	\caption{\label{tab:exp-ewiki-example} The clustering results on $4$ groups of categories. Our method outperforms the others by at least $11.1$ accuracy points.}
\end{table*}

\subsubsection{Analysis}

In this part, we provide a deep insight to the clustering process through experiment on $4$-category problem in Fig.~\ref{fig:exp-20ng-evol}.

In the first graph (graph (a)), we exhibits the evolution of the square error cost (black squares) during the training procedure and the corresponding accuracy value (red empty square) respectively. We find there are two flat parts (denoted with two horizontal grey dotted lines) on the curve inferring some local minimums during parameter update. From these flat parts, we split the training procedure into four regions, marked with I, II, III and IV in Fig.~\ref{fig:exp-20ng-evol}.

The transforming from one region to the next are generally accompanied with the iterative instability which is described in Sec.~\ref{sec:Method-Difficulty}. This instability is reflected by the non-monotonous or dramatic change of the model parameters. We select the mean and max LSTM layer values as an example shown in graph (b) (we rescaled the parameter value curve for better visualization).

In graph (c), we show the detailed clustering results at $4$ stages. For each stage, we have four clusters and each cluster contains instances with different true labels denoted by different colors. At the beginning stage (I) in training, parameters have not been well updated and `comp' dominates $3$ of the $4$ clusters we predicted. Now we can predict only $2$ types of category named `comp' and `talk'. In the second stage (II), the changes happen in the second cluster (from the left) and we are able to predict three types of category.  In stage III, our model enters into the final way to correctly organize all clusters. Each cluster is dominated by instances with different category. At last (stage IV), the distributions are further optimized in all clusters.

\subsection{English Wiki}

\subsubsection{Dataset}

We downloaded the corpus from the English wiki website \footnote{https://en.wikipedia.org/wiki/Main\_Page/}. We remove the structural information (including the head part, tail part, $etc.$) from webpages, and only keep plain texts in the main body. Each sentence is considered as a single instance. Wee keep the original sentence order in the corpus. There are $40$ million sentences in this corpus with vocabulary size of $4$ million.

\subsubsection{Model Training}

We assume that two sentences next to each other have the same topic. On the contrary, two sentences far from each other have different topics with high probability. Following this assumption, we build the positive pair by selecting the consecutive two sentences. We build the negative pair by randomly selecting two sentences, the distance between which is greater than $100$ sentences. The model topology is the same as shown in Fig.~\ref{fig:method-model-unsup} and the hyperparameters are the same as those used in previous two corpus. We cluster this corpus in $100$ clusters.

\subsubsection{Results and analysis}

\begin{figure}[!htb]
	\begin{center}
		\includegraphics[angle=0,width=0.8\columnwidth]{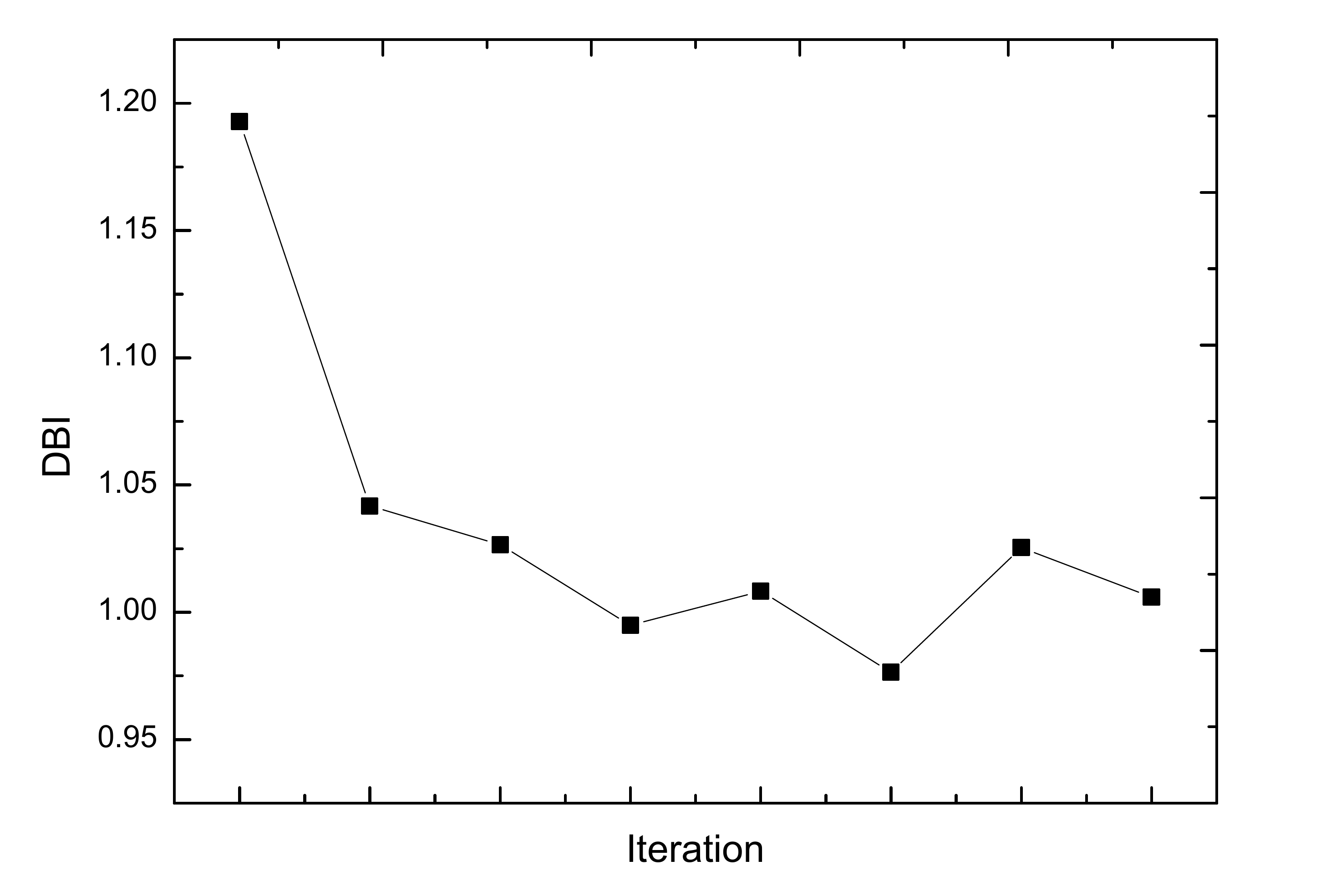}
		\caption{DBI curve during the training process.}
		\label{fig:Wiki-DBI}
	\end{center}
\end{figure}

Conventional clustering tools generally are not able to deal with such a large corpus. We only exhibits the performance of our model. There is no ground truth for exact evaluation. We show the intrinsic metrics DBI in Fig.~\ref{fig:Wiki-DBI}. DBI measures the ratio of cluster radius over the distance between cluster centroid in the worst case. The value $1$ denotes the sum of two cluster radius is equivalent to their distance, which means clusters are separated.

We show the example clusters in Tab.~\ref{tab:exp-ewiki-example}. Both cluster-1 and cluster-5 describe the games. Cluster-1 focus on sports games while cluster-5 focus on electronic games. Meanwhile, sentence in cluster-5 also refers to weapons (see the last sentence), but it can be distinguished from the cluster-6 about war related topic. Furthermore, sentences with rare overlap words can also be clustered together, reflecting the advantage of the purely neural based end-to-end system.


\section{Related Work}

Conventional text clustering methods are mainly based on Expectation-Maximization (EM) algorithms like k-Means \cite{manning2008introduction}. However, k-Means can only give the hard boundaries among clusters. The distance between each instance and its centroid cannot be naturally converted to the probabilistic distribution. This property also results in a difficulty for it to be leveraged by downstream tools. Furthermore, its performance relies on its initialization. Latent Dirichlet Allocation (LDA) \cite{Blei-Jordan-JMLR2003} is an unsupervised method that clusters similar words into topic groups. LDA assumes the multinomial distribution of each word and the corresponding parameters are drawn from the Dirichlet distribution. However, for large corpus, the information distribution may deviate from the assumed distributions.

The basics of text clustering is measuring the similarity of two texts, which is the distance between two text representations. Traditional text representations like bag-of-words and term frequency-inverse document frequency (TF-IDF) cause sparsity problems for short texts. The dense vector representation of text can be constructed by the ensemble of word embeddings \cite{Mikolov2013Efficient} in the text. The Siamese CBOW model \cite{kenter2016siamese} constructs the sentence embedding by averaging the word embeddings and uses the embedding similarities among the sentence, its adjacent sentences and randomly chosen sentences as training target to fine tune the sentence embedding. The Word Mover's Distance (WMD) \cite{kusner2015word} measures the similarity of two texts by calculating the minimum accumulate distance from all the embedded words in one text to the embedded words in the other text. These methods ignore the syntax of words order and the semantic relationship of texts.

Several semantic representation of text (sequence embedding) methods based on neural networks have been proposed and showed advantages in a variety of downstream natural language understanding tasks. The Paragraph Vector (PV) \cite{Le-Mikolov-ICML2014} learns the text embedding by leveraging the text representation as context to predict following word using the paragraph vector and word vectors together. The Skip-Thought Vector (STV) model \cite{kiros2015skip,tang2017trimming} is an encoder-decoder neural network that learns the sequence embedding directly by predicting the surrounding sequences of each input sequence. Hill $et~al.$ \shortcite{hill2016learning} provides systematic evaluation and comparison of unsupervised models that construct distributed representations of texts. However, the optimal text representation method depends on different tasks.

Recently, models that directly learn pairwise text similarities are proposed based on siamese networks \cite{bromley1994signature}. Siamese convolutional neural networks followed by similarity measurement layer are constructed by He $et~al.$ \shortcite{he2015multi} to learn text semantic representations and trained with similarity labeled text pairs. Mueller $et~al.$ \shortcite{mueller2016siamese} present a Siamese LSTM network that scores the similarity of two sentences. The similarity is calculated with the Manhattan distance between text representations. However, to train these models, sequence pairs with well labeled similarity scores are required. The Deep Structured Semantic Model (DSSM) \cite{huang2013learning} has a siamese like structure that learns the query phrase embedding and the document embedding in the common semantic space with deep neural networks, using the cosine similarity between the representations of queries and documents as the target. Inspired by these works, we attempt to employ the siamese deep neural network for end-to-end text clustering. We utilize unlabeled corpus and construct training instances with adjacent and distant sequences pairs. Rather than scoring the similarity of text representations, we target the similarity of the category distributions generated from the two sequence representations in a text pair.

\section{Conclusion}

We present a purely neural based end-to-end method for unsupervised text clustering. The sequence representation learning and clustering model are integrated in an unified framework. We evaluated our model on two widely used benchmarks, IMDB movie reviews and $20$-Newsgroup. The clustering results outperform the other methods by a large margin on both tasks. Our model exhibits the strong ability in resistance to the data noise introduced by our pseudo labels. It exhibits even better performance when we address tasks with more category such as $20$-newsgroup.

Under this framework, there still exist several aspects to improve the model further due to its flexibility. More sampling strategies might be explored to construct instances with higher confidence. We can also change this single pair topology to a pair-wise topology taking double pairs of instances as input. That is, the model takes two pairs of sequences as input for each time, and determines if one pair is inclined to be positive than the other. At last and of the most importance, we expect this end-to-end property could contribute to a wide range of complex natural language understanding tasks.

\bibliography{Main}
\bibliographystyle{acl_natbib}

\end{document}